\documentclass[final]{cvpr}

\usepackage{times}
\usepackage{epsfig}
\usepackage{graphicx}
\usepackage{amsmath}
\usepackage{amssymb}

\usepackage[pagebackref=true,breaklinks=true,colorlinks,bookmarks=false]{hyperref}

\def\confYear{CVPR 2021}

\begin{document}

\title{Bag of Tricks for Neural Architecture Search}

\author{Thomas Elsken$^{1}$, Benedikt Staffler$^1$, Arber Zela$^{2}$, Jan Hendrik Metzen$^1$ and Frank Hutter$^{2,1}$\\
$^1$Bosch Center for Artificial Intelligence, $^2$University of Freiburg\\
{\tt\small   $\{$thomas.elsken, benediktsebastian.staffler, janhendrik.metzen$\}$@de.bosch.com}  \\ {\tt \small  $\{$zelaa, fh$\}$@cs.uni-freiburg.de} 
}
\maketitle

\begin{abstract}

While neural architecture search methods have been successful in previous years and led to new state-of-the-art performance on various problems, they have also been criticized for being unstable, being highly sensitive with respect to their hyperparameters, and often not performing better than random search. To shed some light on this issue, we discuss some practical considerations that help improve the stability, efficiency and overall performance.

\end{abstract}

 
\section{Introduction}
\label{sec:practical}

Neural architecture search (NAS) methods have been very successful in previous years and led to a new state of the art on various problems and benchmarks, e.g., for image classification~\cite{zoph-arXiv18,real_regularized_2018}, semantic segmentation~\cite{Liu_2019_CVPR} or object detection~\cite{Guo_2020_CVPR}; please refer to the surveys~\cite{elsken_survey,Wistuba2019ASO} for an overview. However, they have also been criticized for being unstable and for providing unfair or non-transparent empirical comparisons due to using various tweaks for boosting performance beside just comparing the optimized architecture, see, e.g., \cite{best_practices}. In particular methods employing one-shot models 
have been reported to be brittle, highly sensitive with respect to their hyperparameters, and often no better than random search~\cite{li_repro_nas,sciuto19, Xu_2019_ICCV, Zela2020Understanding, Zela2020NAS-Bench-1Shot1:, Yang2020NAS}. Tricks for stabilizing the search are often hidden in the details and are hard to find for the reader, or are not even discussed.

In this short paper, we provide some insights in the details of using NAS methods and discuss common practices in NAS that help improving the stability (Section \ref{sec:stabilize}), efficiency (Section \ref{sec:speed}), and overall performance (Section \ref{sec:boost}). 

\section{Stabilizing Gradient-Based NAS and Training One-Shot Models}
\label{sec:stabilize}

\paragraph{Weights Warm-Up.} Gradient-based NAS methods typically employ a continuous relaxation of the architecture search space by considering a weighted combination of operations (such as convolution or pooling layers)~\cite{darts}. This allows to search for architectures by using alternating stochastic gradient descent, which (in each batch) iterates updates of the network parameters and the real-valued weights parameterizing the architecture. 
However, directly using this alternating optimization has been reported to lead to premature convergence in the architectural space~\cite{Liu_2019_CVPR}. Consequently, a common trick~\cite{Liu_2019_CVPR,Saikia_2019_ICCV,NIPS2019_9576,Xu_2019_ICCV,Elsken20,Guo_2020_CVPR} is to start by optimizing the network weights only, often for as long as half of the overall search epochs; architecture updates are only conducted afterwards. This trick is important in order for the architecture search to not favour architectures that train faster (in particular those that contain many skip connections).

Similar approaches for warming up weights can be found for sampling-based methods. Bender et al.~\cite{bender_icml:2018} start by training the whole one-shot model and then drop out more and more paths over the course of training. TuNAS~\cite{Bender_2020_CVPR} adapts this strategy; while they directly samples paths from the one-shot models for training, they enable all operations within a certain block of the one-shot model rather than only the sampled operation. The probability for enabling all operations is annealing to 0 over the course of training. 

It is even a possibility to first \emph{fully} train the one-shot model and conduct the search afterwards, thus decoupling these two stages~\cite{bender_icml:2018,guo2020single,NIPS2019_8890}.

\paragraph{Regularization and Loss Landscape Smoothing.} 
It was shown~\cite{Zela2020Understanding} that smoothing the loss landscape by using stronger regularization can help to stabilize architecture search. This can, e.g., be done via stochastic regularization techniques, such as drop path~\cite{zoph-arXiv18}, weight decay or data augmentation. Alternatively, more robust loss functions can achieve a similar goal, e.g., by minimizing the loss in a neighbourhood of an optimal architecture rather than only for the optimum~\cite{chen2020stabilizing} or by implicitly smoothing the loss function via additional auxiliary connections~\cite{chu2021darts}.

\paragraph{Normalization layers.}  For NAS methods using a continuous relaxation of the search space, such as DARTS, a naive use of normalization layers such as batch~\cite{Ioff15}, layer~\cite{layer_norm}, instance~\cite{instance_norm} or group~\cite{Wu_2018_ECCV} normalization, is problematic since their learnable parameters can lead to a rescaling of the architectural parameters and thus make them meaningless. Consequently, the learnable parameters are typically disabled~\cite{darts}. Xu et al.~\cite{Xu_2019_ICCV} even report that in general batch normalization was harmful in their experiments and hence they do not use it at all. Furthermore, batch normalization can cause issues in combination with NAS methods that require to keep the one-shot model in memory since this naturally leads to using small batch sizes due to memory limitations. 
This is especially problematic for applications with high-resolution input images.

Some normalization layers are also fundamentally problematic in combination with sampling-based methods since the normalization statistics will vary across different sampled paths. Bender et al.~\cite{bender_icml:2018} report that training the one-shot models was highly unstable in early stages of experimentation, and that these instabilities were overcome by using batch statistics also during evaluation and a variant of ghost batch normalization~\cite{NIPS2017_6770}. 
Many researchers also replace standard batch normalization by more advanced techniques, e.g., \cite{NIPS2019_8890} use synchronized batch normalization~\cite{peng2018megdet} across GPUs to increase the effective batch size and recalculate batch statistics during architecture optimization and \cite{Wang_2020_CVPR} use group normalization~\cite{Wu_2018_ECCV} instead. We also refer to \cite{understanding_sna} for a discussion of batch normalization within models trained by sampling paths.

\section{Speeding Up NAS}
\label{sec:speed}

\paragraph{Proxy Tasks.}
A very common approach for speeding up NAS is to use lower fidelity (or proxy) estimates. E.g.,  approaches using a cell-based search space typically use fewer cells with fewer filters during search than during evaluation~\cite{zoph-arXiv18} and train for fewer epochs. 
The size of the trainig data set can also be reduced to make the search more efficient,
e.g., by downscaling images~\cite{Liu_2019_CVPR} or by searching on a smaller data set (e.g., CIFAR or PennTreeBank)
and transferring the learned cells to a larger one (e.g. ImageNet or WikiText-2) as is often done in practice~\cite{zoph-arXiv18, real_regularized_2018, darts}.

We refer to Elsken et al.~\cite{elsken_survey} for a general overview. Zhou et al.~\cite{Zhou_2020_CVPR} study the impact of such lower fidelity estimates and assess how different proxies should be used in combination to achieve the best speed up while maintaining a high correlation with the true optimization metric.

\paragraph{Feature Caching.} Recently, many researchers have applied NAS methods to tasks such as semantic segmentation~\cite{Liu_2019_CVPR} or object detection~\cite{Xu_2019_ICCV}, where architectures are composed of several components, such as a backbone and a task-specific head. When the backbone is fixed during the search, its outputs can be \emph{pre-computed} once for all training data points to avoid unnecessary computation and thereby speed up architecture search~\cite{NIPS2018_8087, Nekrasov_2019_CVPR, Wang_2020_CVPR}.

 \paragraph{Speeding Up The Optimization Process via Sequential Search.} Rather than optimizing different components of an architecture jointly, the search is often split up into several phases for different components in order to reduce memory and time consumption. For example, in the case of object detection, Xu et al.~\cite{Xu_2019_ICCV} first search for the multi-scale feature extractor and then for the detection head.
Du et al.~\cite{Du_2020_CVPR} first search for scale permutations of a given network and then tune the building blocks of the resulting architecture, e.g., by adjusting the resolution of feature maps and by choosing one out of a set of predefined possible building blocks, such as a residual block or a bottleneck block.

 Guo et al.~\cite{Guo_2020_CVPR} first sequentially screen different search spaces for different architectural components with a downscaled model and prune the search spaces before conducting a final optimization of the reduced search spaces.

\paragraph{Pre-Optimized Search Spaces.}

While in principle NAS can be viewed as a subfield of automated machine learning (AutoML)~\cite{automl_book} and thus aims for searching for architectures with as little prior knowledge from humans as possible, it can nevertheless be helpful to build search spaces around architectures that are known to work well for efficiency reasons, rather than searching from scratch~\cite{real2020automl}. For example, search spaces are often based on inverted
residual blocks~\cite{Sand18}, essentially resulting in optimizing the hyperparameters that come with these blocks, such as kernel sizes, expansion ratios or dilatation rates~\cite{Shaw_2019_ICCV,Guo_2020_CVPR,Bender_2020_CVPR,Chen_2020_CVPR}.
Some methods also directly build upon existing architectures and search for transformations of these architectures, e.g., via permuting layers~\cite{Du_2020_CVPR} or by searching how to connect channel groups within an architecture~\cite{NIPS2019_9576}.
We note that while this use of pre-optimized search spaces is likely to yield improved results for a particular application more quickly, this process cannot discover entirely new architectures, such as Transformer~\cite{vaswani_2017_transformer} architectures.
In order to achieve the latter, one would have to use dramatically more powerful search spaces, and potentially with a hierarchical structure~\cite{Liu17, Xie_2019_ICCV, ru_2020_generator}.

\section{Improving the Final Performance}
\label{sec:boost}

 \paragraph{Deriving Optimal Architectures from the Search Process.} Identifying the optimal architectures from NAS runs is not trivial for at least the following reasons: firstly, as almost all methods employ lower fidelity estimates, the ranking of architectures on the proxy tasks will likely be different from the ranking on the true task. Secondly, it is currently not well understood how weight sharing affects the ranking of architectures. Some researchers show that weight sharing is not necessarily properly ranking architectures~\cite{li_repro_nas,sciuto19,Yang2020NAS,Zela2020NAS-Bench-1Shot1:}. 

 Consequently, researchers often first collect a set of candidate architectures, either by running NAS multiple times~\cite{darts} or by obtaining multiple architectures from a single run of the method (e.g., by sampling from a learned distribution or by sampling from a population of evolved networks)~\cite{zoph-arXiv18,NIPS2018_8087}. These sets of candidate architectures are then evaluated in a setting which has higher correlation with respect to the setting of interest and the best out of the candidates is chosen to be the optimal architecture. This process is sometimes also already used within the search process when components are searched sequentially~\cite{Wang_2020_CVPR}.
 To increase correlation between ranking of architectures with weights inherited from the one-shot model versus when retrained from scratch, Zhao et al.~\cite{zhao2020fewshot} propose to use a set of sub-one-shot models, where each sub model covers different regions of the search space, with the goal of alleviating undesired co-adaption.
 Additionally, for approaches employing a continuous relaxation of the search space, it remains unclear what the best way is to obtain a discretized architecture from the real-valued parameterization.

 Typically, the operations with maximum weight are chosen as initially proposed by Liu et al.~\cite{darts}. Wang et al. \cite{wang2021rethinking} argue that this process is suboptimal since the operation weights are not directly correlated with performance of the resulting architecture and thus propose a different scheme for extracting a discretized architecture based on minimizing the drop in performance when removing an operation from the one-shot model.

\paragraph{Hyperparameters, Data Augmentation and other Tweaks for Boosting Performance.}

The performance of a neural architecture depends on many factors other than the architecture itself, such as data augmentation~\cite{DeVr17,Zhan18d,Cubuk_2019_CVPR}, stochastic regularization~\cite{DBLP:journals/corr/Gastaldi17,zoph-arXiv18}, activation functions~\cite{ramachandran2018searching} and other hyperparameters such as learning rate (schedules)~\cite{DBLP:journals/corr/LoshchilovH16a}. Yang et al.~\cite{Yang2020NAS} provide a thorough ablation study on these factors on CIFAR-10.
They show that the training pipeline is more important than the architecture: The worst out of eight randomly sampled architectures trained with the best training pipeline substantially outperformed the best of the eight architectures using the worst training pipeline.
To give another example, MobileNetV3~\cite{Howard_2019_ICCV} achieved $75.2\%$ top-1 accuracy on ImageNet, suggesting an improvement of $3.2\%$ due to the novel architecture compared to the performance of $72.0\%$ for MobileNetV2~\cite{Sand18}. 
However, Bender et al.~\cite{Bender_2020_CVPR} show that when both models are trained with an identical state-of-the-art training pipeline, MobileNetV2 achieves $73.3\%$ accuracy compared to $75.3\%$ for MobileNetV3, thereby reducing the improvement due to the architecture from $3.2\%$ to $2.0\%$. Thus, all these factors along with the architecture heavily impact the final performance.
Moreover, the search hyperparameters are in particular important for one-shot NAS methods as already discussed above. 
Zela et al.~\cite{Zela2020NAS-Bench-1Shot1:} optimize the hyperparameters of various one-shot NAS algorithms and show that the found solutions can outperform black-box NAS optimizers when properly tuned. 
To avoid many of these confounding factors when comparing different NAS algorithms, a series of NAS benchmarks~\cite{ying2019_nasbench,Zela2020NAS-Bench-1Shot1:,Dong2020NAS-Bench-201:,nb301,mehrotra2021nasbenchasr,li2021hwnasbench} have been proposed.

\section{Conclusion}

We presented a list of tips and tricks for employing NAS methods and making them more robust in practice. We hope that these can ease the usability of NAS methods, both for experienced and new researchers.

\bibliography{main}

\end{document}